\def\year{2021}\relax
\documentclass[letterpaper]{article} 
\usepackage{aaai21}  
\usepackage{times}  
\usepackage{helvet} 
\usepackage{courier}  
\usepackage[hyphens]{url}  
\usepackage{graphicx} 
\usepackage{natbib}  
\usepackage{caption} 
\usepackage{amssymb}
\usepackage{multirow}
\usepackage[dvipsnames]{xcolor}
\usepackage{colortbl} 

\definecolor{mygray}{gray}{0.6}
\definecolor{mygray-bg}{gray}{0.9}

\newcommand{\ie}{\textit{i}.\textit{e}.}
\newcommand{\eg}{\textit{e}.\textit{g}.}
\newcommand{\cf}{\textit{cf.}}

\usepackage{pdfrender}

\usepackage{bm}
\usepackage{amsmath}
\usepackage{subfig}
\usepackage[switch]{lineno}

\title{Ref-NMS: \\ Breaking Proposal Bottlenecks in Two-Stage Referring Expression Grounding}

\author {
	Long Chen,\textsuperscript{\rm 2}\thanks{indicates equal contribution (\{longc, mwb\}@zju.edu.cn).} Wenbo Ma,\textsuperscript{\rm 1}\footnotemark[1] Jun Xiao,\textsuperscript{\rm 1}\thanks{indicates corresponding author (junx@zju.edu.cn).} Hanwang Zhang,\textsuperscript{\rm 3} Shih-Fu Chang\textsuperscript{\rm 4} \\
}

\affiliations {
	\textsuperscript{\rm 1} College of Computer Science, Zhejiang University, Hangzhou \\ \textsuperscript{\rm 2} Tencent AI Lab, Shenzhen \\
	\textsuperscript{\rm 3} MReaL Lab, Nanyang Technological University, Singapore \\\textsuperscript{\rm 4} DVMM Lab, Columbia University, New York \\
}

\begin{document}
	

\maketitle

\begin{abstract}
	The prevailing framework for solving referring expression grounding is based on a two-stage process: 1) detecting proposals with an object detector and 2) grounding the referent to one of the proposals. Existing two-stage solutions mostly focus on the grounding step, which aims to align the expressions with the proposals. In this paper, we argue that these methods overlook an obvious \emph{mismatch} between the roles of proposals in the two stages: they generate proposals solely based on the detection confidence (\ie, expression-agnostic), hoping that the proposals contain all right instances in the expression (\ie, expression-aware). Due to this mismatch, current two-stage methods suffer from a severe performance drop between detected and ground-truth proposals. To this end, we propose Ref-NMS, which is the first method to yield expression-aware proposals at the first stage. Ref-NMS regards all nouns in the expression as critical objects, and introduces a lightweight module to predict a score for aligning each box with a critical object. These scores can guide the NMS operation to filter out the boxes irrelevant to the expression, increasing the recall of critical objects, resulting in a significantly improved grounding performance. Since Ref-NMS is agnostic to the grounding step, it can be easily integrated into any state-of-the-art two-stage method. Extensive ablation studies on several backbones, benchmarks, and tasks consistently demonstrate the superiority of Ref-NMS. Codes are available at: \url{https://github.com/ChopinSharp/ref-nms}.
\end{abstract}

\section{Introduction}

Referring Expression Grounding (REG), \ie, localizing the targeted instance (referent) in an image given a natural language description, is a longstanding task for multimodal understanding. Considering different granularities of localization, there are two sub-types of REG: 1) \textbf{Referring Expression Comprehension (REC)}~\cite{hu2017modeling,hu2016natural,yu2016modeling,yu2017joint}, where the referents are localized by bounding boxes (bboxes). 2) \textbf{Referring Expression Segmentation (RES)}~\cite{hu2016segmentation,liu2017recurrent,shi2018key,margffoy2018dynamic}, where the referents are localized by segmentation masks. Both two tasks are important for many downstream high-level applications such as VQA~\cite{antol2015vqa}, navigation~\cite{chen2019touchdown}, and autonomous driving~\cite{kim2019grounding}.

\begin{figure}[t]
	\centering
	\includegraphics[width=0.8\linewidth]{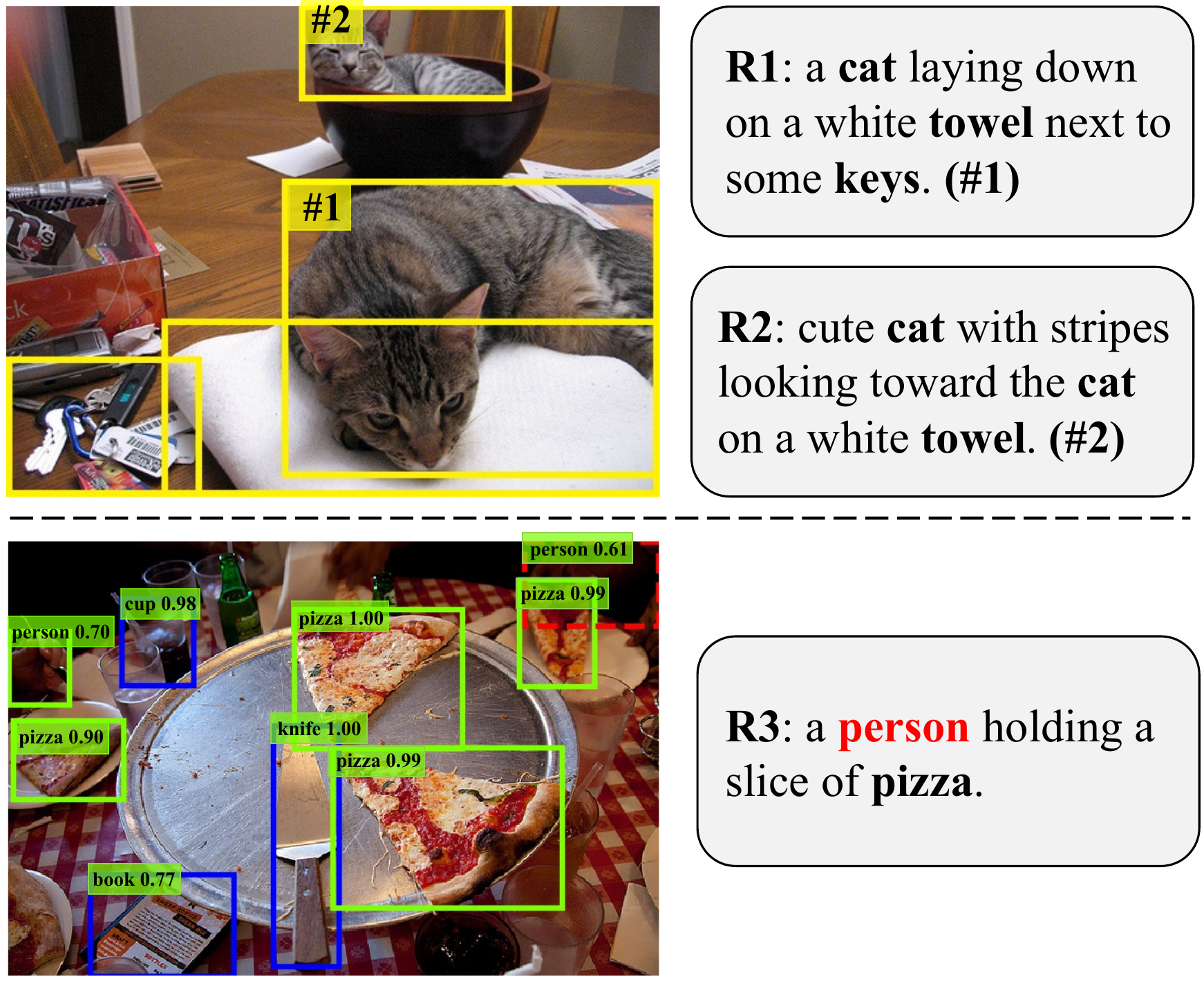}
	\caption{Upper: A typical REC example from RefCOCOg. The two ``similar" expressions (R1 and R2) refer to different objects. Below: An example of proposals from the first-stage of prevailing MAttNet~\cite{yu2018mattnet}. The proposals only contain bboxes with high detection confidence ($>$ 0.65) regardless of the content of expression (\eg, The candidates \texttt{knife}, \texttt{book}, and \texttt{cup} are not mentioned in R3). Red dashline bbox denotes the missing referent.}
	\label{fig:motivation_ac}
\end{figure}

State-of-the-art REG methods can be classified into two major categories: one-stage, proposal-free methods and two-stage, proposal-driven methods. For the one-stage methods~\cite{chen2018real,yang2019fast,liao2020real}, they regard REG as a generalized object detection (or segmentation) task, and the whole textual expression is treated as a specific object category. Although these one-stage methods achieve faster inference speed, their grounding performance, especially for complex expressions (\eg, in dataset RefCOCOg), is still behind the two-stage counterpart. The main reasons for the differences are two-fold: 1) The one-stage methods naturally focus on the local content, \ie, they fail to perform well in the expressions which need global reasoning. For example in Figure~\ref{fig:motivation_ac}, when grounding ``\emph{a cat laying down on a white towel next to some keys}", it is even difficult for humans to identify the referent \texttt{cat} without considering its contextual objects \texttt{towel} and \texttt{keys}. 2) The one-stage methods do not exploit the linguistic structure of expressions, \ie, they are not sensitive to linguistic variations in expressions. For instance, when changing the expression in Figure~\ref{fig:motivation_ac} to ``\emph{cute cat with stripes looking toward the cat on a white towel}", they tend to refer to the same object (\#1)~\cite{akula2020words}. On the contrary, the two-stage methods~\cite{yu2018mattnet,liu2019learning,liu2019improving} intuitively are more similar to the human way of reasoning: 1) detecting proposals with a detector, and then 2) grounding the referent to one of the proposals. In general, two-stage methods with perfect proposals (\eg, all human-annotated object regions) can achieve more accurate and explainable grounding results than the one-stage methods. 

\begin{figure}[t]
	\centering
	\includegraphics[width=0.7\linewidth]{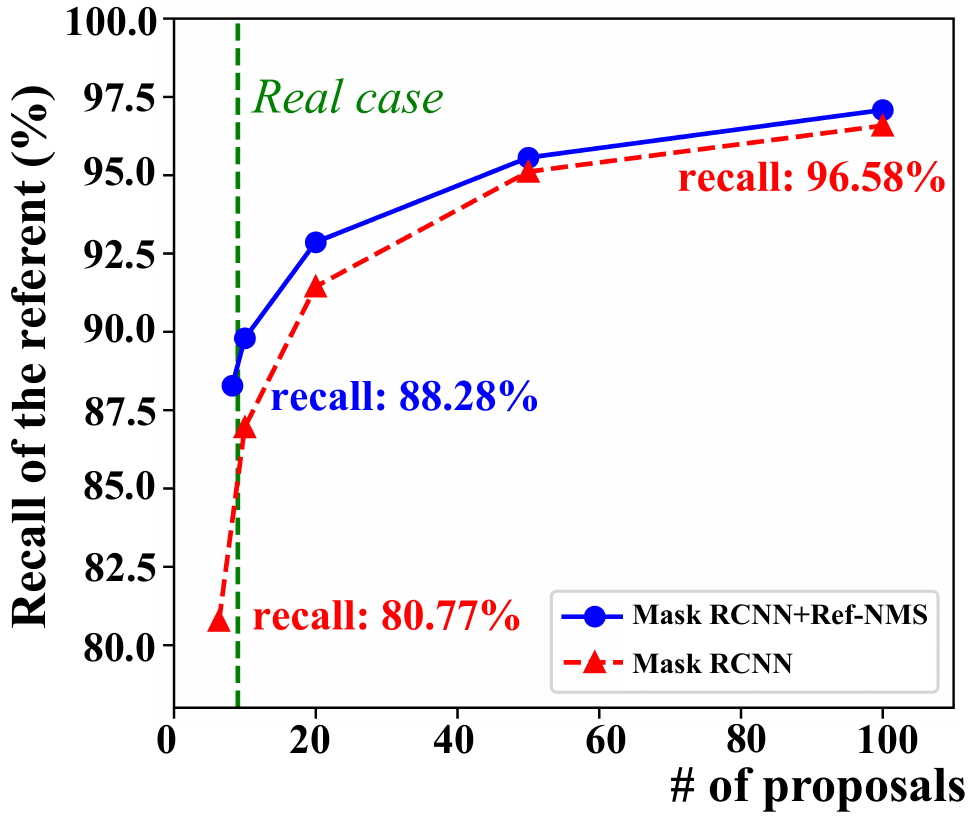}
	\caption{The recall of the referent (IoU$>$0.5) vs. number of proposals on the RefCOCO testB set. The real case denotes the actual situation in all SOTA two-stage methods.}
	\label{fig:motivation_b}
\end{figure}

Unfortunately, when using the results from off-the-shelf detectors as proposals, two-stage methods' performance all drops dramatically. This is also the main weakness of two-stage solutions often criticized by competing methods in the literature, \ie, the performance of two-stage methods is heavily limited by the proposal quality. In this paper, we argue that this huge performance gap between the detected and ground-truth proposals is mainly caused by the \textbf{mismatch} between the roles of proposals in the two stages: \emph{the first-stage network generates proposals solely based on the detection confidence, while the second-stage network just assumes that the generated proposals will contain all right instances in the expression}. More specifically, for each image, a well pre-trained detector can detect hundreds of detections with a near-perfect recall of the referent and contextual objects (\eg, as shown in Figure~\ref{fig:motivation_b}, recall of the referent can reach up to 96.58\% with top-100 detections). However, to relieve the burden of the referent grounding step in the second stage, current two-stage methods always filter proposals simply based on their detection confidences. These heuristic rules result in a sharp reduction of the recall (\eg, decrease to 80.77\% as in Figure~\ref{fig:motivation_b}), and bring in the mismatch negligently. To illustrate this further, we show a concrete example in Figure~\ref{fig:motivation_ac}. To ground the referent at the second stage, we hope that the proposals contain the referent \texttt{person} and its contextual object \texttt{pizza}. In contrast, the first-stage network only keeps bboxes with high detection confidence (\eg, \texttt{knife}, \texttt{book}, and \texttt{cup}) as proposals, but actually misses the critical referent \texttt{person} (\ie, the red bbox).

In this paper, we propose a novel algorithm Ref-NMS, to rectify the mismatch of detected proposals at the conventional first stage. In particular, for each expression, Ref-NMS regards all nouns in the expression as critical objects, and introduces a lightweight relatedness module to predict a probability score for each proposal to be a critical object. The higher predicted score denotes the higher relevance between a proposal and the expression. Then, we fuse the relatedness scores and classification scores, and exploit the fused scores as the suppression criterion in Non-Maximum Suppression (NMS). After NMS, we can filter out the proposals with little relevance to the expression. Finally, all proposals and the expression are fed into the second-stage grounding network, to obtain the referent prediction.

We demonstrate the significant performance gains of Ref-NMS on three challenging REG benchmarks. It's worth noting that the Ref-NMS can be generalized and easily integrated into any state-of-the-art two-stage method to further boost its performance on both REC and RES. Our method is robust and efficient, opening the door for many downstream applications such as multimodal summarization.

\begin{figure*}[t]
	\centering
	\includegraphics[width=0.8\linewidth]{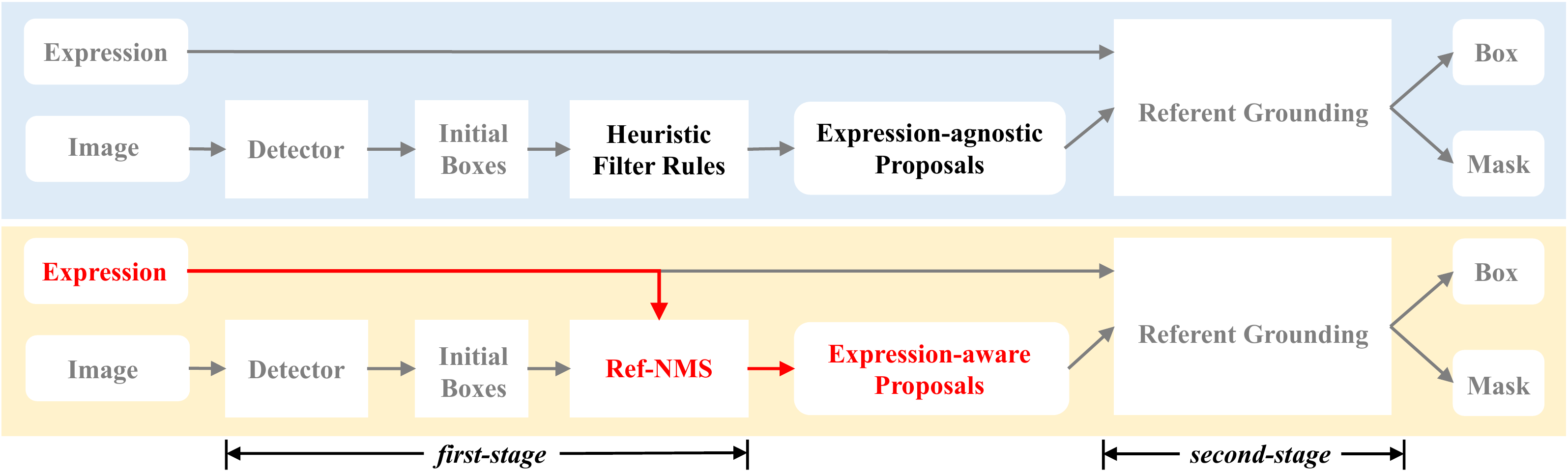}
	\caption{Upper: A typical two-stage REG framework, which uses heuristic filter rules to obtain expression-agnostic proposals at the first-stage, and feeds them into the second stage for referent grounding. Below: The Ref-NMS module can generate expression-aware proposals by considering the expression at the first stage.}
	\label{fig:two-stage-models}
\end{figure*}


\section{Related Work}

\textbf{Referring Expression Comprehension (REC).} Current overwhelming majority of REC methods are in a two-stage manner: proposal generation and referent grounding. To the best of our knowledge, existing two-stage works all focus on the second stage. Specifically, they tend to design a more explainable reasoning process by structural modeling~\cite{yu2018mattnet,liu2019improving,liu2019learning,liu2019joint,hong2019learning,niu2019variational}, or more effective multi-modal interaction mechanism~\cite{wang2019neighbourhood,yang2020graph}. However, their performance is strictly limited by the proposals from the first stage. Recently, another emerging direction to solve REC is in a one-stage manner~\cite{chen2018real,yang2019fast,liao2020real,luo2020multi,yang2020improving}. Although one-stage methods achieve faster inference speed empirically, they come at a cost of lost interpretability and poor performance in composite expressions. In this paper, we rectify the overlooked mismatch in two-stage methods.

\textbf{Referring Expression Segmentation (RES).} Unlike REC, most of RES works are one-stage methods. They typically utilize a "concatenation-convolution" design to combine the two different modalities: they first concatenate the expression feature with visual features at each location, and then use several conv-layers to fuse the multimodal features for mask generation. To further improve mask qualities, they usually enhance their backbones with more effective features by multi-scale feature fusion~\cite{margffoy2018dynamic}, feature progressive refinement~\cite{li2018referring,chen2019see,huang2020referring}, or novel attention mechanisms~\cite{shi2018key,ye2019cross,hu2020bi}. Besides, with the development of two-stage instance segmentation (\eg, Mask R-CNN~\cite{he2017mask}), two-stage REC methods can be extended to solve RES simply by replacing the object detection network at the second stage to an instance segmentation network. Analogously, Ref-NMS can be easily integrated into any two-stage RES method.

\textbf{Phrase Grounding}. It is a task closely related to REC. There are also two types of solutions: proposal-free and proposal-driven methods. Different from REC, the queries in phrase grounding have two characteristics: 1) Simple. This relieves two-stage methods from complicated relational reasoning and allow them to accept more proposals (\eg, $>$ 200 proposals)\footnote{In contrast, the average number of proposals in REC is 10.} at the second stage, which means two-stage phrase grounding methods doesn't suffer from the aforementioned recall drop problem. 2) Diverse. Efforts have been taken to address this problem by either using a object detector pre-trained on another large-scale dataset~\cite{yu2018rethinking} or re-generate proposals with respect to queries and mentioned objects~\cite{chen2017query}.

\textbf{Non-Maximum Suppression (NMS).} NMS is a de facto standard post-processing step adopted by numerous modern object detectors, which removes duplicate bboxes based on detection confidence. Except for the most prevalent GreedyNMS, multiple improved variants have been proposed recently. Generally, they can be categorized into three groups: 1) Criterion-based~\cite{jiang2018acquisition,tychsen2018improving,tan2019learning,yang2019learning}: they utilize other scores instead of classification confidence as the criterion to remove bboxes by NMS, \eg, IoU scores. 2) Learning-based~\cite{hosang2017learning,hu2018relation}: they directly learn an extra network to remove duplicate bboxes. 3) Heuristic-based~\cite{bodla2017soft,liu2019adaptive}: they dynamically adjust the thresholds for suppression according to some heuristic rules. In this paper, we are inspired by the criterion-based NMS, and design the Ref-NMS, which uses both expression relatedness and detection confidence as the criterion.

\section{Approach}


\subsection{Revisiting Two-Stage REG Framework}\label{sec:revisit}
The two-stage framework is the most prevalent pipeline for REG. As shown in Figure~\ref{fig:two-stage-models}, it consists of two separate stages: proposal generation at the first-stage and referent grounding at the second-stage.

\textbf{Proposal Generation.} Given an image, current two-stage methods always resort to a well pre-trained detector to obtain a set of initially detected bboxes, and utilize an NMS to remove duplicate bboxes. However, even after NMS operation, there are still thousands of bboxes left (\eg, each image in RefCOCO has an average of 3,500 detections). To relieve the burden of the following referent grounding step, all existing works further filter these bboxes based on their detection confidences. Although this heuristic filter rule can reduce the number of proposals, it also results in a drastic drop in the recall of both the referent and contextual objects (Detailed results are reported in Table~\ref{tab:hit-rate}.).

\textbf{Referent Grounding.} In the training phase, two-stage methods usually use the ground-truth regions in COCO as proposals, and the number is quite small (\eg, each image in RefCOCO has an average of 9.84 ground-truth regions). For explainable grounding, state-of-the-art two-stage methods always compose these proposals into graph~\cite{yang2019dynamic,wang2019neighbourhood} or tree~\cite{liu2019learning,hong2019learning} structures, \ie, as the number of proposals increases linearly, the number of computation increases exponentially. Therefore, in the test phase, it is a must for them to filter detections at the first stage.

\begin{figure*}[t]
	\centering
	\includegraphics[width=0.8\linewidth]{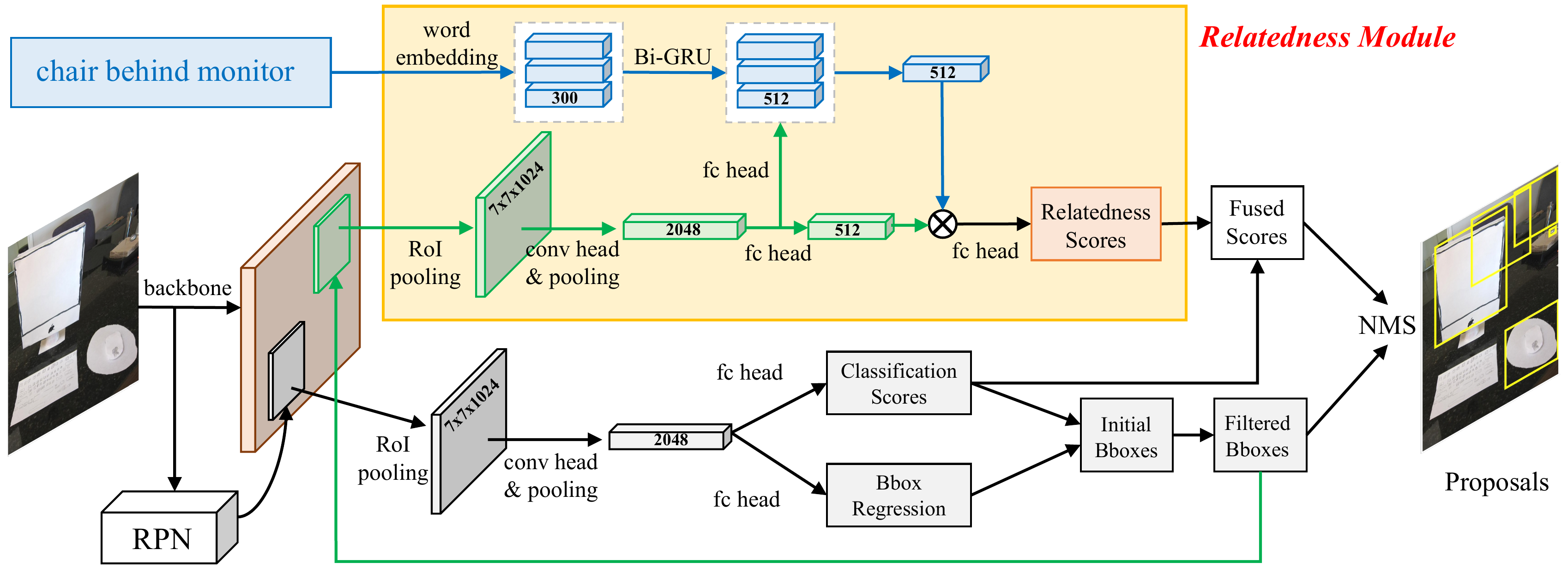}
	\caption{The overview of Ref-NMS model. Given an image, the model uses a pre-trained detector to generate thousands of initial bboxes. Then, hundreds of filtered bboxes and the expression are fed into the relatedness module to predict their relatedness scores. Lastly, the fused scores are used as the suppression criterion of NMS.} 
	\label{fig:architecture}
\end{figure*}

\subsection{Relatedness Module}

An overview of the Ref-NMS model is shown in Figure~\ref{fig:architecture}. The core of Ref-NMS is the relatedness module. Given an image and a pre-trained detector, we can receive thousands of initial bboxes. To reduce the computation of the relatedness module, we first use a threshold $\delta$ to filter the bboxes with classification confidence, and obtain a filtered bbox set $\mathcal{B}$. For each bbox $\bm{b}_i \in \mathcal{B}$, we use a region visual encoder $e_v$ (\ie, an RoI Pooling layer and a convolutional head network) to extract the bbox feature $\bm{v}_i \in \mathbb{R}^v$. Meanwhile, for the referring expression $Q$, we use an expression encoder $e_q$ (\ie, a Bi-GRU) to output a set of word features $\{\bm{w}_1, ..., \bm{w}_{|Q|}\}$, where $\bm{w}_j \in \mathbb{R}^q$ is the $j$-th word feature. For each bbox $\bm{b}_i$, we use a soft-attention mechanism~\cite{chen2017sca} to calculate a unique expression feature $\bm{q}_i$ by:
\begin{equation} \label{eq:1}
\begin{aligned} 
\bm{v}^a_i = \text{MLP}_a (\bm{v}_i), \quad & a_{ij} = \text{FC}_s ([\bm{v}^a_i; \bm{w}_j]), \\
\alpha_{ij} = \text{softmax}_j(a_{ij}),  \quad & \bm{q}_i = \textstyle{\sum_j} \alpha_{ij} \bm{w}_j,
\end{aligned}
\end{equation}
where $\text{MLP}_a$ is a two-layer MLP mapping $\bm{v}_i \in \mathbb{R}^v$ to $\bm{v}^a_i \in \mathbb{R}^q$, $\text{FC}_s$ is a FC layer to calculate the similarity between bbox feature $\bm{v}^a_i$ and word feature $\bm{w}_j$, and $[;]$ is a concatenation operation. Then, we combine the two modal features and predict the relatedness score $r_i$:
\begin{equation}
\begin{aligned}
\bm{v}^b_i = \text{MLP}_b(\bm{v}_i), \quad & \bm{m}_i = \text{L2Norm}(\bm{v}^b_i \odot \bm{q}_i), \\
\hat{r}_i = \text{FC}_r(\bm{m}_i), \quad & r_i = \text{sigmoid}(\hat{r}_i),
\end{aligned}
\end{equation}
where $\text{MLP}_b$ is a two-layer MLP mapping $\bm{v}_i \in \mathbb{R}^v$ to $\bm{v}^b_i \in \mathbb{R}^q$, $\odot$ is the element-wise multiplication, L2Norm represents $l_2$ normalization, and $\text{FC}_r$ is a FC layer mapping $\bm{m}_i \in \mathbb{R}^q$ to $\hat{r}_i \in \mathbb{R}$.

\textbf{Score Fusion.} After obtaining the relatedness score $r_i$ for bbox $\bm{b}_i$, we multiply $r_i$ with the classification confidence $c_i$ for bbox $\bm{b}_i$ from the original detector, and utilize the multiplication of two scores $s_i$ as the suppression criterion of the NMS operation, \ie, $s_i = r_i \times c_i$.

\subsection{Training Objectives for Ref-NMS}

To learn the relatedness score for each bbox, we need the ground-truth annotations for all mentioned instances (\ie, both referent and contextual objects) in the expression. However, current REG datasets only have annotations about the referent. Thus, we need to generate pseudo ground-truths for contextual objects. Specifically, we first assign POS tags to each word in the expression using the spaCy POS tagger and extract all nouns in the expression. Then, we calculate the cosine similarity between GloVe embeddings of extracted nouns and categories of ground-truth regions in COCO\footnote{Two-stage methods always use an object detector pretrained on COCO dataset. Thus, we don't use extra or more annotations.}. Lastly, we use threshold $\gamma$ to filter regions as the pseudo ground-truths. 

In the training phase, we regard all the pseudo ground-truth bboxes and annotated referent bboxes as foreground bboxes. And we use two types of training objectives:

\textbf{Binary XE Loss.} For each bbox $\bm{b}_i \in \mathcal{B}$, if it has a high overlap (\ie, IoU$>$0.5) with any foreground bbox, its ground-truth relatedness score $r^*$ is set to 1, otherwise $r^*=0$. Then the relatedness score prediction becomes a binary classification problem. We can use the binary cross-entropy (XE) loss as the training objective:
\begin{equation}
\begin{aligned}
L = -\frac{1}{|\mathcal{B}|} \textstyle{\sum^{|\mathcal{B}|}_{i=1}} r^*_i \log(r_i) + (1-r^*_i) \log(1-r_i).
\end{aligned}
\end{equation}

\textbf{Ranking Loss.} Generally, if a bbox has a higher IoU with foregound bboxes, the relatedness between the bbox and expression should be higher, \ie, we can use the ranking loss as the training objectives: 
\begin{equation}
L = \frac{1}{N} \textstyle{\sum_{(\bm{b}_i, \bm{b}_j), \rho_i < \rho_j}} \max (0, r_i - r_j + \alpha),
\end{equation}
where $\rho_i$ denotes the largest IoU value between bbox $\bm{b}_i$ and foreground bboxes, $N$ is the total number of pos-neg training pairs, and $\alpha$ is a constant to control the ranking margin, set as 0.1. To select the pos-neg pair $(\bm{b}_i, \bm{b}_j)$, we follow the sampling-after-splitting strategy~\cite{tan2019learning}. Specifically, we first divide the bbox set $\mathcal{B}$ into 6 subsets based on a quantization $q$-value: $q_i = \lceil \text{max}(0, \rho_i - 0.5)/0.1  \rceil$, \ie, the bbox with higher IoU value has larger $q$-value. Then, all bboxes with $\rho > 0.5$ are selected as positive samples. For each positive sample, we rank the top-$h$ bboxes as negative samples based on predicted relatedness scores from the union of subsets with smaller $q$-value.

\addtolength{\tabcolsep}{-3pt}
\begin{table*}[t]
	\small
	\begin{center}
		\scalebox{0.88}{
			\begin{tabular}{| c| c | c c c | c c c | c c || c c c | c c c | c c |}
				\hline
				& \multirow{3}{*}{Ref-NMS} & \multicolumn{8}{c||}{Referent} & \multicolumn{8}{c|}{Contextual Objects} \\
				\cline{3-18}
				& & \multicolumn{3}{c|}{RefCOCO} & \multicolumn{3}{c|}{RefCOCO+} & \multicolumn{2}{c||}{RefCOCOg} & \multicolumn{3}{c|}{RefCOCO} & \multicolumn{3}{c|}{RefCOCO+} & \multicolumn{2}{c|}{RefCOCOg} \\
				& & val & testA & testB &  val & testA & testB & val & test & val & testA & testB &  val & testA & testB & val & test \\
				\hline
				\parbox[t]{2mm}{\multirow{3}{*}{\rotatebox[origin=c]{90}{N=100}}} &  & 97.60 & 97.81 & 96.58 & 97.79 & 97.78 & 96.99 & 97.18 & 96.91 & 90.14 & 89.85 & 90.53 & 89.53 & 88.47 & 90.69 & 90.56 & 90.30 \\
				& B & 97.75 & 98.59 & 97.08 & 97.96 & 98.39 & 97.50 & 97.61 & 97.44 & 90.38 & 90.31 & 90.64 & 89.67 & 88.88 & 91.04 & 90.36 & 90.37  \\
				& R & 97.62 & 98.02 & 96.78 & 97.71 & 98.06 & 97.14 & 97.18 & 97.08 & 90.22 & 89.83 & 90.63 & 89.70 & 88.62 & 90.71 & 90.67 & 90.30  \\
				\hline
				\parbox[t]{2mm}{\multirow{3}{*}{\rotatebox[origin=c]{90}{Real}}} &  & 88.84 & 93.99 & 80.77 & 90.71 & 94.34 & 84.11 & 87.83 & 87.88 & 74.97 & 78.60 & 70.19 & 76.34 & 77.45 & 73.52 & 75.69 & 75.87 \\
				& B & \textbf{92.51} & \textbf{95.56} & \textbf{88.28} & \textbf{93.42} & \textbf{95.86} & \textbf{88.95} & \textbf{90.28} & \textbf{90.34} & \textbf{78.75} & \textbf{80.14} & \textbf{76.47} & \textbf{78.44} & \textbf{78.82} & \textbf{77.49} & 76.12 & 76.57  \\
				& R & 90.50 & 94.75 & 83.87 & 91.62 & 95.14 & 86.42 & 89.01 & 88.96 & 76.79 & 79.12 & 72.99 & 77.66 & 78.44 & 75.59 & \textbf{76.68} & \textbf{76.73}  \\
				\hline
			\end{tabular}
		} 
	\end{center}
	\caption{Recall (\%) of the referent and contextual objects. The baseline detector is the ResNet-101 based Mask R-CNN with plain GreedyNMS. B denotes the Ref-NMS with binary XE loss, R denotes the Ref-NMS with ranking loss. Real denotes the real case used in the state-of-the-art two-stage methods.}
	\label{tab:hit-rate}
\end{table*}
\addtolength{\tabcolsep}{3pt}

\addtolength{\tabcolsep}{-3pt}    
\begin{table*}[t]
	\small
	\begin{center}
		\scalebox{0.88}{
			\begin{tabular}{| l | c c c | c c c | c c || c c c | c c c | c c |}
				\hline
				\multirow{3}{*}{Models} & \multicolumn{8}{c||}{Referring Expression Comprehension} & \multicolumn{8}{c|}{Referring Expression Segmentation} \\
				\cline{2-17}
				& \multicolumn{3}{c|}{RefCOCO} & \multicolumn{3}{c|}{RefCOCO+} & \multicolumn{2}{c||}{RefCOCOg} & \multicolumn{3}{c|}{RefCOCO} & \multicolumn{3}{c|}{RefCOCO+} & \multicolumn{2}{c|}{RefCOCOg} \\
				& val & testA & testB &  val & testA & testB & val & test & val & testA & testB &  val & testA & testB & val & test \\
				\hline
				MAttNet~\cite{yu2018mattnet}  & 76.65 & 81.14 & 69.99 & 65.33 & 71.62 & 56.02 & 66.58 & 67.27 & 56.51 & 62.37 & 51.70 & 46.67 & 52.39 & 40.08 & 47.64 & 48.61 \\
				MAttNet$^\dagger$ & 76.92 & 81.19 & 69.58 & 65.90 & \textbf{71.53} & 57.23 & 67.52 & 67.55 & 57.14  &  62.34  & 51.48  & 47.30  & \textbf{52.37} & 41.14 & 48.28  & 49.01 \\
				~~+Ref-NMS B & \textbf{78.82} & \textbf{82.71} & \textbf{73.94} & \textbf{66.95} & 71.29 & \textbf{58.40} & \textit{68.89} & \textbf{68.67} & \textbf{59.75} & \textbf{63.48} & \textbf{55.66} & \textbf{48.39} & 51.57 & \textbf{42.56} & \textit{49.54} & \textbf{50.38} \\
				~~+Ref-NMS R & \textit{77.98} & \textit{82.02} & \textit{71.64} & \textit{66.64} & \textit{71.36} & \textit{58.01} & \textbf{69.16} & \textit{67.63} & \textit{58.32} & \textit{62.96} & \textit{53.68} & \textit{47.87} & \textit{51.85} & \textit{41.41} & \textbf{50.13} & \textit{49.07} \\
				\hline
				NMTree~\cite{liu2019learning} & 76.41 & 81.21 & 70.09 & 66.46 & 72.02 & 57.52 & 65.87 & 66.44 & 56.59 & 63.02 & 52.06 & 47.40 & 53.01 & 41.56 & 46.59 & 47.88 \\
				NMTree$^\dagger$ & 76.54 & 81.32 & 69.66 & 66.65 & 71.48 & 57.74 & 65.65 & 65.94 & 56.99 & \textit{62.88} & 51.90 & 47.75 & \textit{52.36} & 41.86 & 46.19 & 47.41 \\
				~~+Ref-NMS B & \textbf{78.67} & \textbf{82.09} & \textbf{73.78} & \textbf{67.15} & \textit{71.76} & \textit{58.70} & \textbf{67.30} & \textbf{66.93} & \textbf{59.95} & \textbf{63.25} & \textbf{55.64} & \textbf{48.68} & 52.30 & \textbf{42.64} & \textbf{48.14} & \textbf{48.59} \\
				~~+Ref-NMS R &  \textit{77.81} & \textit{81.69} & \textit{71.78} & \textit{67.03} & \textbf{71.78} & \textbf{58.79} & \textit{66.81} & \textit{66.31} & \textit{58.42} & 62.69 & \textit{53.60} & \textit{48.27} & \textbf{52.65} & \textit{42.18} & \textit{47.72} & \textit{48.09} \\
				\hline
				CM-A-E~\cite{liu2019improving}  & 78.35 & 83.14 & 71.32 & 68.09 & 73.65 &	58.03 & 67.99 & 68.67  & --- & --- & --- & --- & --- & --- & --- & --- \\
				CM-A-E$^\dagger$ & 78.35 & 83.12 & 71.32 & 68.19 & 73.04 & 58.27 & 69.10 & 69.20 & 58.23  & 64.60 & 53.14  & 49.65  & \textbf{53.90}  & 41.77  & 49.10  & 50.72 \\
				~~+Ref-NMS B & \textbf{80.70} & \textbf{84.00} & \textbf{76.04} & \textit{68.25} & \textbf{73.68} & \textbf{59.42} & \textbf{70.55} & \textbf{70.62} & \textbf{61.46} & \textbf{65.55} & \textbf{57.41} & \textit{49.76} & \textit{53.84} & \textbf{42.66} & \textbf{51.21} & \textbf{51.90}  \\
				~~+Ref-NMS R & \textit{79.55} & \textit{83.58} & \textit{73.62} & \textbf{68.51} & \textit{73.14} & \textit{58.38} & \textit{69.77} & \textit{70.01} & \textit{59.72} & \textit{64.87} & \textit{55.63} & \textbf{49.86} & 52.62 & \textit{41.87} & \textit{50.13} & \textit{51.44} \\
				\hline
			\end{tabular}
		} 
	\end{center}
	\caption{Performances of different architectures on REC and RES. The metrics are top-1 accuracy (\%) for REC and overall IoU (\%) for RES. All baselines use the ResNet-101 based Mask R-CNN as first-stage networks. The best and second best methods under each setting are marked in bold and italic fonts, respectively. $^\dagger$ denotes the results from our implementations.}
	\label{tab:model-agnostic}
\end{table*}
\addtolength{\tabcolsep}{3pt} 

\section{Experiments}
\subsection{Experimental Settings and Details}

\textbf{Datasets.} We evaluate the Ref-NMS on three challenging REG benchmarks: 1) \textbf{RefCOCO}~\cite{yu2016modeling}:  It consists of 142,210 referring expressions for 50,000 objects in 19,994 images. These expressions are collected in an interactive game interface~\cite{kazemzadeh2014referitgame}, and the average length of each expression is 3.5 words. All expression-referent pairs are split into train, val, testA, and testB sets. The testA set contains the images with multiple people and the testB set contains the images with multiple objects.
2) \textbf{RefCOCO+}~\cite{yu2016modeling}: It consists of 141,564 referring expressions for 49,856 objects in 19,992 images. Similar to RefCOCO, these expressions are collected from the same game interface, and have train, val, testA, and testB splits. 
3) \textbf{RefCOCOg}~\cite{mao2016generation}: It consists of 104,560 referring expressions for 54,822 objects in 26,711 images. These expressions are collected in a non-interactive way, and the average length of each expression is 8.4 words. We follow the same split as~\cite{nagaraja2016modeling}.

\textbf{Evaluation Metrics.} For the REC task, we use the top-1 accuracy as evaluation metric. When the IoU between bbox and ground truth is larger than 0.5, the prediction is correct. For the RES task, we use the overall IoU and Pr@X (the percentage of samples with IoU higher than X)\footnote{Due to the limited space, all RES results with the Pr@X metric are provided in the supplementary materials.} as metrics.

\textbf{Implementation Details.} We build a vocabulary for each dataset by filtering the words less than 2 times, and exploit the 300-d GloVe embeddings as the initialization of word embeddings. We use an "unk" symbol to replace all words out of the vocabulary. The largest length of sentences is set to 10 for RefCOCO and RefCOCO+, 20 for RefCOCOg. The hidden size of the encoder $e_q$ is set to 256. For encoder $e_v$, we use the same head network of the Mask R-CNN with ResNet-101 backbone\footnote{https://github.com/lichengunc/mask-faster-rcnn} as prior works~\cite{yu2018mattnet}, and utilize the pre-trained weights as initialization. The weights of the original detector (\ie, the gray part in Figure~\ref{fig:architecture}) are fixed during training. The whole model is trained with Adam optimizer. The learning rate is initialized to 4e-4 and 5e-3 for the head network and the rest of network. We set the batch size as 8. The thresholds $\delta$ and $\gamma$ are set to 0.05 and 0.4, respectively. For ranking loss, the top-h is set to 100.

\begin{table*}[t]
	\small
	\begin{center}
		\scalebox{0.88}{
			\begin{tabular}{|l | l | l | c | c c c | c c c | c c|}
				\hline
				& \multirow{2}{*}{Models} & \multirow{2}{*}{Venue} & \multirow{2}{*}{Backbone} & \multicolumn{3}{c|}{RefCOCO} & \multicolumn{3}{c|}{RefCOCO+} & \multicolumn{2}{c|}{RefCOCOg}  \\
				&  &  &  & val & testA & testB & val & testA & testB & val & test \\
				\hline
				\parbox[t]{2mm}{\multirow{4}{*}{\rotatebox[origin=c]{90}{one-s.}}} & SSG~\cite{chen2018real} & \textit{arXiv'18} &  darknet53 & --- & 76.51 & 67.50 & --- & 62.14 & 49.27 & 58.80 & --- \\  
				& FAOA~\cite{yang2019fast} & \textit{ICCV'19} & darknet53 & 71.15 & 74.88 & 66.32 & 56.86 & 61.89 & 49.46 & 59.44 & 58.90 \\
				& RCCF~\cite{liao2020real} & \textit{CVPR'20} & dla34 & --- & 81.06 & 71.85 & --- & 70.35 & 56.32 & --- & 65.73 \\
				& RSC-Large~\cite{yang2020improving} & \textit{ECCV'20} & darknet53 & 77.63 & 80.45 & 72.30 & 63.59 & 68.36 & 56.81 & 67.30 & 67.20 \\
				\hline
				\hline
				\parbox[t]{2mm}{\multirow{10}{*}{\rotatebox[origin=c]{90}{two-s.}}} & VC~\cite{zhang2018grounding} & \textit{CVPR'18} & vgg16 & --- & 73.33 & 67.44 & --- & 58.40 & 53.18 & --- & --- \\
				& ParalAttn~\cite{zhuang2018parallel} & \textit{CVPR'18} & vgg16 & --- & 75.31 & 65.52 & --- & 61.34 & 50.86 & --- & --- \\
				& LGRANs~\cite{wang2019neighbourhood} & \textit{CVPR'19} & vgg16 & --- & 76.60 & 66.40 & --- & 64.00 & 53.40 & --- & --- \\
				& DGA~\cite{yang2019dynamic} & \textit{ICCV'19} & vgg16 & --- & 78.42 & 65.53 & --- & 69.07 & 51.99 & --- & 63.28 \\ 
				& NMTree~\cite{liu2019learning} & \textit{ICCV'19} & vgg16 & 71.65 & 74.81 & 67.34 & 58.00 & 61.09 & 53.45 & 61.01 & 61.46 \\
				& MAttNet~\cite{yu2018mattnet} & \textit{CVPR'18} &	res101 & 76.65 & 81.14 & 69.99 & 65.33 & 71.62 & 56.02 & 66.58 & 67.27 \\
				& RvG-Tree~\cite{hong2019learning} & \textit{TPAMI'19} & res101 & 75.06 & 78.61 & 69.85 & 63.51 & 67.45 & 56.66 & 66.95 & 66.51 \\
				& NMTree~\cite{liu2019learning} & \textit{ICCV'19} & res101 & 76.41 & 81.21 & 70.09 & 66.46 & 72.02 & 57.52 & 65.87 & 66.44 \\
				& CM-A-E~\cite{liu2019improving} & \textit{CVPR'19}	& res101 &	78.35 &	83.14 & 71.32 &	68.09 &	73.65 &	58.03 & 67.99 & 68.67 \\ 
				& \cellcolor{mygray-bg}{\textbf{CM-A-E+Ref-NMS}} & \cellcolor{mygray-bg}{\textit{AAAI'21}} & \cellcolor{mygray-bg}{res101} &
				\cellcolor{mygray-bg}{\textbf{80.70}} & \cellcolor{mygray-bg}{\textbf{84.00}} & \cellcolor{mygray-bg}{\textbf{76.04}} & \cellcolor{mygray-bg}{\textbf{68.25}} & \cellcolor{mygray-bg}{\textbf{73.68}} & \cellcolor{mygray-bg}{\textbf{59.42}} & \cellcolor{mygray-bg}{\textbf{70.55}} & \cellcolor{mygray-bg}{\textbf{70.62}} \\
				\hline
			\end{tabular}
		} 
	\end{center}
	\caption{Top-1 accuracies (\%) of state-of-the-art models on referring expression comprehension.}
	\label{tab:sota_rec}
\end{table*}

\begin{table*}[t]
	\small
	\begin{center}
		\scalebox{0.88}{
			\begin{tabular}{| l | l | l | c c c | c c c | c c |}
				\hline
				& \multirow{2}{*}{Models} & \multirow{2}{*}{Venue} & \multicolumn{3}{c|}{RefCOCO} & \multicolumn{3}{c|}{RefCOCO+} & \multicolumn{2}{c|}{RefCOCOg}  \\
				& & & val & testA & testB & val & testA & testB &  val & test \\
				\hline
				\parbox[t]{2mm}{\multirow{4}{*}{\rotatebox[origin=c]{90}{one-s.}}} 
				& STEP~\cite{chen2019see} & \textit{ICCV'19} & 60.04 & 63.46 & 57.97 & 48.19 & 52.33 & 40.41 & --- & --- \\
				& BRINet~\cite{hu2020bi} & \textit{CVPR'20} & 60.98 & 62.99 & 59.21 & 48.17 & 52.32 & 42.41 & --- & --- \\
				& CMPC~\cite{huang2020referring} & \textit{CVPR'20} &  61.36 & 64.53 & 59.64 & 49.56 & 53.44 & 43.23 & --- & --- \\
				& MCN~\cite{luo2020multi} & \textit{CVPR'20} & 62.44 & 64.20 & 59.71 & 50.62 & 54.99 & 44.69 & 49.22 & 49.40 \\
				\hline
				\hline
				\parbox[t]{2mm}{\multirow{4}{*}{\rotatebox[origin=c]{90}{two-s.}}}  & MAttNet~\cite{yu2018mattnet} & \textit{CVPR'18} & 56.51 & 62.37 & 51.70 & 46.67 & 52.39 & 40.08 & 47.64 & 48.61 \\
				& NMTree~\cite{liu2019learning} & \textit{ICCV'19} & 56.59 & 63.02 & 52.06 & 47.40 & 53.01 & 41.56 & 46.59 & 47.88 \\
				& CM-A-E$^\dagger$~\cite{liu2019improving} & \textit{CVPR'19} & 58.23  & 64.60 & 53.14  & 49.65  & \textbf{53.90}  & 41.77  & 49.10  & 50.72 \\
				& \cellcolor{mygray-bg}{\textbf{CM-A-E+Ref-NMS}} & \cellcolor{mygray-bg}{\textit{AAAI'21}} & \cellcolor{mygray-bg}{\textbf{61.46}} & \cellcolor{mygray-bg}{\textbf{65.55}} & \cellcolor{mygray-bg}{\textbf{57.41}} & \cellcolor{mygray-bg}{\textbf{49.76}} & \cellcolor{mygray-bg}{53.84} & \cellcolor{mygray-bg}{\textbf{42.66}} & \cellcolor{mygray-bg}{\textbf{51.21}} & \cellcolor{mygray-bg}{\textbf{51.90}} \\ 
				\hline
			\end{tabular}
		} 
	\end{center}
	\caption{Overall IoU (\%) of state-of-the-art models on referring expression segmentation. All methods utilize ResNet-101 as backbone. $^\dagger$ denotes that the results are from our reimplementation. Note that since one-stage RES and two-stage RES models are always pretrained on different datasets, the comparision between one-stage and two-stage models are not absolutely fair.}
	\label{tab:sota_res}
\end{table*}

\subsection{Recall Analyses of Critical Objects}

\textbf{Settings.} To evaluate the effectiveness of the Ref-NMS to improve the recall of both referent and contextual objects, we compare Ref-NMS with plain GreedyNMS used in the baseline detector (\ie, ResNet-101 based Mask R-CNN). Since we only have annotated ground-truth bboxes for the referent, we calculate the recall of pseudo ground-truths to approximate the recall of contextual objects. The results are reported in Table~\ref{tab:hit-rate}, and more detailed results are provided in the supplementary materials.

\textbf{Results.} From Table~\ref{tab:hit-rate}, we have the following observations. When using top-100 bboxes as proposals, all three methods can achieve near-perfect recall ($\approx$ 97\%) for the referent and acceptable recall ($\approx$ 90\%) for the contextual objects, respectively. However, when the number of proposals decreases to a very small number (\eg, $<$ 10 in the real case), the recall of the baseline all drops significantly (\eg, 15.81\% for the referent and 20.34\% for the contextual objects on RefCOCO testB). In contrast, Ref-NMS can help narrow the gap over all dataset splits. Especially, the improvement is more obvious in the testB set (\eg, 7.51\% and 4.85\% absolute gains for the recall of referent on RefCOCO and RefCOCO+), where the categories of referents are more diverse and the recalls are relatively lower.

\begin{figure*}[t]
	\centering
	\includegraphics[width=0.93\linewidth]{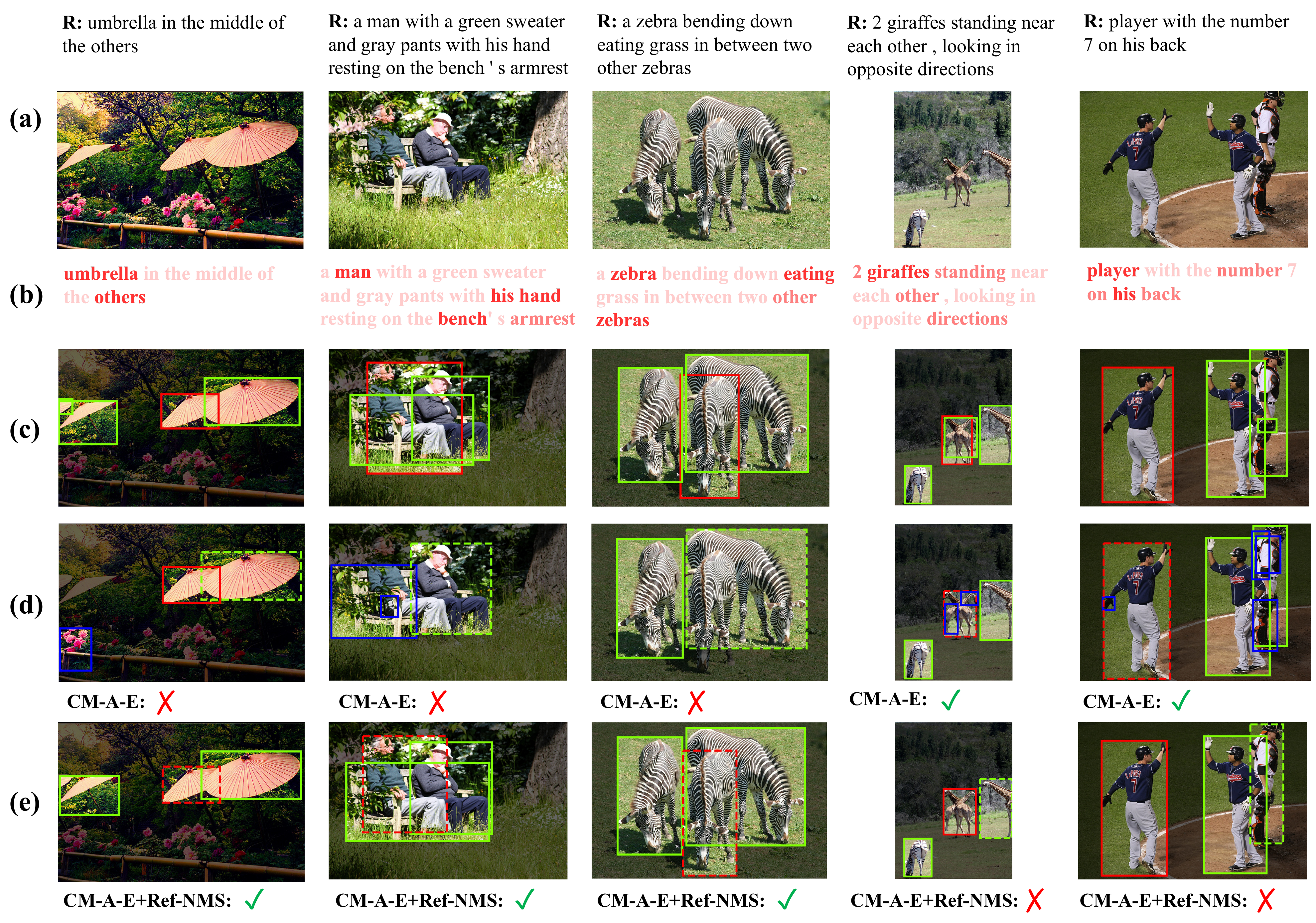}
	\caption{Qualitative REC results on RefCOCOg showing comparisons between correct (green tick) and wrong referent grounds (red cross) by CM-A-E and CM-A-E+Ref-NMS. (a): The input image and referring expressions. (b): The visualiation of word attention weights $\alpha$ (\cf, Eq.~\eqref{eq:1}) for each referent object. (c): The annotated referent ground-truth bbox (red) and generated pseudo ground-truth bboxes for contextual objects (green). (d) and (e) denote the proposals and final grounding results from two methods. We only show the proposals and the final predicted referent bbox is illustrated in dash line. The denotations of bbox colors are as follows. Red: The bbox hits (IoU$>$0.5) the referent ground-truth bbox; Green: The bboxes hit the pseudo ground-truth bboxes; Blue: The false positive proposal predictions.}
	\label{fig:visualization}
\end{figure*}

\subsection{Architecture Agnostic Generalization}

\textbf{Settings.} Since the Ref-NMS model is agnostic to the second stage network, it can be easily integrated into any referent grounding architectures. To evaluate the effectiveness and generality of Ref-NMS to boost the grounding performance of different backbones, we incorporated the Ref-NMS into multiple SOTA two-stage methods: \textbf{MAttNet}~\cite{yu2018mattnet} , \textbf{NMTree}~\cite{liu2019learning}, and \textbf{CM-A-E}~\cite{liu2019improving}. All results are reported in Table~\ref{tab:model-agnostic}.

\textbf{Results.} From Table~\ref{tab:model-agnostic}, we can observe that both variants of Ref-NMS can consistently improve the grounding performance over all three backbones on both REC and RES. The improvement is more significant on the testB set (\eg, 4.72\% and 3.23\% absolute performance gains for CM-A-E in REC and RES), which meets our expectation, \ie, the improvements of grounding performance have a strong positive correlation with the improvements of the recall of critical objects. Compared between two variants of Ref-NMS, in most of cases, Ref-NMS B achieves better grounding performance. We argue that the reason may come from the imbalance of positive and negative samples in each level.

\subsection{Comparisons with State-of-the-Arts}

We incorporate Ref-NMS (with binary XE loss) into model CM-A-E, which is dubbed \textbf{CM-A-E+Ref-NMS}, and compare it against the state-of-the-art REC and RES methods.

\textbf{Settings.} For the state-of-the art REC methods, from the viewpoint of one-stage and two-stage, we can group them into: 1) Two-stage methods: \textbf{VC}~\cite{zhang2018grounding}, \textbf{ParalAttn}~\cite{zhuang2018parallel},\textbf{LGRANs}~\cite{wang2019neighbourhood}, \textbf{DGA}~\cite{yang2019dynamic}, \textbf{NMTree}~\cite{liu2019learning}, \textbf{MAttNet}~\cite{yu2018mattnet}, \textbf{RvG-Tree}~\cite{hong2019learning}, and \textbf{CM-A-E}~\cite{liu2019improving}; 2) one-stage methods: \textbf{SSG}~\cite{chen2018real}, \textbf{FAOA}~\cite{yang2019fast}, \textbf{RCCF}~\cite{liao2020real}, and \textbf{RSC-Large}~\cite{yang2020improving}. Analogously, for the state-of-the-art RES methods, we group them into: 1) Two-stage methods: \textbf{MAttNet}~\cite{yu2018mattnet}, \textbf{NMTree}~\cite{liu2019learning}, and \textbf{CM-A-E}~\cite{liu2019improving}; 2) one-stage methods: 
\textbf{STEP}~\cite{chen2019see}, \textbf{BRINet}~\cite{hu2020bi}, \textbf{CMPC}~\cite{huang2020referring}, and \textbf{MCN}~\cite{luo2020multi}.

\textbf{Results.} The REC and RES results are reported in Table~\ref{tab:sota_rec} and Table~\ref{tab:sota_res}. For the REC, CM-A-E+Ref-NMS achieves a new record-breaking performance that is superior to all existing REC methods on three benchmarks. Ref-NMS improves the strong baseline CM-A-E with an average of 2.64\%, 0.53\%, and 2.26\% absolute performance gains over RefCOCO, RefCOCO+, and RefCOCOg, respectively. For the RES, CM-A-E+Ref-NMS achieves a new state-of-the-art performance of two-stage methods over most of the dataset splits. Similarly, Ref-NMS improves CM-A-E with an average of 2.82\%, 0.31\%, and 1.65\% absolute performance gains over the three datasets.

\subsection{Qualitative Results}

We illustrate the qualitative results between CM-A-E+Ref-NMS  and baseline CM-A-E on REC in Figure~\ref{fig:visualization}. From the results in line (b), we can observe that Ref-NMS can assign high attention weights to words that are more relevant to individual referents (\eg, umbrella, man, and zebra). The results in line (c) show that the generated pseudo ground-truth bboxes can almost contain all contextual objects in the expression, except a few objects whose categories are far different from the categories of COCO (\eg, sweater, armrest, and grass). By comparing the results between line (d) and line (e), we have the following observations: 1) The baseline method always detects more false-positive proposals (\ie, the blue bboxes), and misses some critical objects (\ie, the red and green bboxes). Instead, Ref-NMS helps the model generate more expression-aware proposals. 2) Even for the failed cases in CM-A-E+Ref-NMS (\ie, the last two columns), Ref-NMS still generates more reasonable proposals (\eg, with less false positive proposals), and the grounding errors mainly come from the second stage.

\section{Conclusions and Future Works}

In this paper, we focused on the two-stage referring expression grounding, and discussed the overlooked mismatch problem between the roles of proposals in different stages. Particularly, we proposed a novel approach dubbed Ref-NMS to calibrate this mismatch. Ref-NMS tackles the problem by considering the expression at the first stage, and learns a relatedness score between each detected proposal and the expression. The multiplication of the relatedness scores and classification scores serves as the suppression criterion for the NMS operation. Meanwhile, Ref-NMS is agnostic to the referent grounding step, and can be integrated into any state-of-the-art two-stage method. 
Moving forward, we plan to apply Ref-NMS into other proposal-drive tasks which suffer from the same mismatch issue, \eg, video grounding~\cite{xiao2021boundary,chen2020rethinking}, VQA~\cite{chen2020counterfactual} and scene graph generation~\cite{chen2019counterfactual}.


\section*{Acknowledgments} This work was supported by the National Natural Science Foundation of China (U19B2043, 61976185), Zhejiang Natural Science Foundation (LR19F020002, LZ17F020001), Major Project of Zhejiang Social Science Foundation (21XXJC01ZD), and the Fundamental Research Funds for the Central Universities.

{
\bibliography{aaai21}
}


\newpage
\section*{Appendix}
\appendix
\setcounter{table}{4}
\setcounter{figure}{5}

\begin{table*}[!h]
	\small
	\begin{center}
		\scalebox{0.9}{
			\begin{tabular}{| c| l | c c c | c c c | c c|}
				\hline
				\multirow{2}{*}{\# proposals}& \multirow{2}{*}{Model}& \multicolumn{3}{c|}{RefCOCO} & \multicolumn{3}{c|}{RefCOCO+} & \multicolumn{2}{c|}{RefCOCOg}  \\
				& & val & testA & testB &  val & testA & testB & val & test  \\
				\hline
					\multirow{3}{*}{N=100} & Baseline & 97.60 & 97.81 & 96.58 & 97.79 & 97.78 & 96.99 & 97.18 & 96.91  \\
					& ~~+ Ref-NMS B & \textbf{97.75} & \textbf{98.59} & \textbf{97.08} & \textbf{97.96} & \textbf{98.39} & \textbf{97.50} & \textbf{97.61} & \textbf{97.44} \\
					& ~~+ Ref-NMS R & 97.62 & 98.02 & 96.78 & 97.71 & 98.06 & 97.14 & 97.18 & 97.08   \\
				\hline
					\multirow{3}{*}{N=50} & Baseline & 96.63 & 97.14 & 95.11 & 96.89 & 97.15 & 95.56 & 95.98 & 95.53 \\
					& ~~+ Ref-NMS B & \textbf{97.16} & \textbf{97.91} & \textbf{95.56} & \textbf{97.23} & \textbf{97.73} & \textbf{96.26} & \textbf{96.49} & \textbf{96.20} \\
					& ~~+ Ref-NMS R & 96.58 & 97.21 & 95.51 & 96.74 & 97.22 & 96.13 & 96.14 & 95.67 \\
				\hline
					\multirow{3}{*}{N=20} & Baseline & 94.54 & 95.85 & 91.44 & 94.81 & 95.90 & 91.92 & 93.42 & 92.96 \\
					& ~~+ Ref-NMS B & \textbf{95.63} & \textbf{96.89} & \textbf{92.86} & \textbf{95.53} & \textbf{96.65} & \textbf{93.56} & \textbf{94.04} & \textbf{94.25} \\
					& ~~+ Ref-NMS R & 94.69 & 96.32 & 91.93 & 94.92 & 96.37 & 92.66 & 93.85 & 93.19 \\
				\hline
					\multirow{3}{*}{N=10} & Baseline & 91.34 & 93.94 & 86.95 & 91.73 & 94.04 & 87.58 & 89.83 & 89.15 \\
					& ~~+ Ref-NMS B & \textbf{93.09} & \textbf{95.63} & \textbf{89.79} & \textbf{93.36} & \textbf{95.72} & \textbf{89.79} & \textbf{91.36} & \textbf{91.19} \\
					& ~~+ Ref-NMS R & 92.20 & 94.73 & 88.68 & 92.51 & 94.81 & 89.51 & 90.52 & 90.20 \\
				\hline				
					\multirow{3}{*}{Real case} & Baseline & 88.84 & 93.99 & 80.77 & 90.71 & 94.34 & 84.11 & 87.83 & 87.88  \\
					& ~~+ Ref-NMS B & \textbf{92.51} & \textbf{95.56} & \textbf{88.28} & \textbf{93.42} & \textbf{95.86} & \textbf{88.95} & \textbf{90.28} & \textbf{90.34} \\
					& ~~+ Ref-NMS R & 90.50 & 94.75 & 83.87 & 91.62 & 95.14 & 86.42 & 89.01 & 88.96   \\
				\hline
			\end{tabular}
		} 
	\end{center}
	\caption{Recall (\%) of the referent with different numbers of proposals.}
	\label{tab:recall-of-referent}
\end{table*}

This supplementary document is organized as follows:
\begin{itemize}
    \item Section~\ref{sec:s1} reports detailed results about the recall of the referent and contextual objects.
	
	\item Section~\ref{sec:s2} reports detailed referring expression segmentation results in the Pr@X metric.
	
	\item Section~\ref{sec:s3} validates the effectiveness of using pseudo ground truths for contextual objects in the training phase. 
	
	\item Section~\ref{sec:s4} validates the effectiveness of our method using Edgebox proposals instead of Mask-RCNN proposals.
	
	\item Section~\ref{sec:s5} shows more qualitative results on RefCOCO, RefCOCO+ and RefCOCOg.
	
\end{itemize}

\begin{table*}[h]
	\small
	\begin{center}
		\scalebox{1.0}{
			\begin{tabular}{| c| l | c c c | c c c | c c|}
				\hline
				\multirow{2}{*}{\# proposals}& \multirow{2}{*}{Model}& \multicolumn{3}{c|}{RefCOCO} & \multicolumn{3}{c|}{RefCOCO+} & \multicolumn{2}{c|}{RefCOCOg}  \\
				& & val & testA & testB &  val & testA & testB & val & test  \\
				\hline
					\multirow{3}{*}{N=100} & Baseline & 90.14 & 89.85 & 90.53 & 89.53 & 88.47 & 90.69 & 90.56 & 90.30 \\
					& ~~+ Ref-NMS B & \textbf{90.38} & \textbf{90.31} & \textbf{90.64} & 89.67 & \textbf{88.88} & \textbf{91.04} & 90.36 & \textbf{90.37} \\
					& ~~+ Ref-NMS R & 90.22 & 89.83 & 90.63 & \textbf{89.70} & 88.62 & 90.71 & \textbf{90.67} & 90.30 \\
				\hline
					\multirow{3}{*}{N=50} & Baseline & 88.72 & 88.37 & 88.78 & 88.09 & 86.90 & 88.85 & 88.87 & 88.25 \\
					& ~~+ Ref-NMS B & 88.69 & 88.45 & 88.51 & 87.81 & \textbf{87.62} & 88.74 & 88.58 & \textbf{88.36} \\
					& ~~+ Ref-NMS R & \textbf{88.76} & \textbf{88.57} & \textbf{88.81} & \textbf{88.12} & 87.15 & \textbf{89.15} & \textbf{89.10} & 88.34 \\
				\hline
					\multirow{3}{*}{N=20} & Baseline & 84.10 & 84.74 & 84.18 & 83.35 & 82.67 & 84.32 & 84.68 & 84.17 \\
					& ~~+ Ref-NMS B & 84.51 & \textbf{85.39} & 83.81 & 83.69 & \textbf{83.47} & 83.74 & 84.65 & \textbf{84.40} \\
					& ~~+ Ref-NMS R & \textbf{84.55} & 85.01 & \textbf{84.21} & \textbf{83.87} & 83.12 & \textbf{84.73} & \textbf{85.10} & 84.24 \\
				\hline
					\multirow{3}{*}{N=10} & Baseline & 76.10 & 76.56 & 76.79 & 75.06 & 73.76 & 77.02 & 78.39 & 78.37 \\
					& ~~+ Ref-NMS B & \textbf{78.26} & \textbf{80.24} & 77.38 & \textbf{77.32} & \textbf{77.84} & 77.17 & 79.12 & \textbf{79.07} \\
					& ~~+ Ref-NMS R & 77.79 & 78.87 & \textbf{77.57} & 76.75 & 76.30 & \textbf{78.39} & \textbf{79.40} & 79.05 \\
				\hline				
					\multirow{3}{*}{Real case} & Baseline & 74.97 & 78.60 & 70.19 & 76.34 & 77.45 & 73.52 & 75.69 & 75.87 \\
					& ~~+ Ref-NMS B & \textbf{78.75} & \textbf{80.14} & \textbf{76.47} & \textbf{78.44} & \textbf{78.82} & \textbf{77.49} & 76.12 & 76.57 \\
					& ~~+ Ref-NMS R & 76.79 & 79.12 & 72.99 & 77.66 & 78.44 & 75.59 & \textbf{76.68} & \textbf{76.73} \\
				\hline
			\end{tabular}
		} 
	\end{center}
	\caption{Recall (\%) of the contexutal objects with different numbers of proposals.}
	\label{tab:recall-of-contextual}
\end{table*}

\addtolength{\tabcolsep}{-3pt}
\begin{table}[h]
	\small
	\begin{center}
		\scalebox{0.85}{
			\begin{tabular}{| l | c c c | c c c | c c|}
				\hline
				\multirow{2}{*}{Model} & \multicolumn{3}{c|}{RefCOCO} & \multicolumn{3}{c|}{RefCOCO+} & \multicolumn{2}{c|}{RefCOCOg}  \\
				& val & testA & testB &  val & testA & testB & val & test  \\
				\hline
					MAttNet & \multirow{2}{*}{11.51} & \multirow{2}{*}{11.37} & \multirow{2}{*}{12.17} & \multirow{2}{*}{10.51} & \multirow{2}{*}{10.20} & \multirow{2}{*}{10.70} & \multirow{2}{*}{11.60} & \multirow{2}{*}{10.91} \\
					~~~~+ Edge-Conf & & & & & & & & \\
					MAttNet & \multirow{2}{*}{\textbf{67.18}} & \multirow{2}{*}{\textbf{70.94}} & \multirow{2}{*}{\textbf{61.98}} & \multirow{2}{*}{\textbf{55.34}} & \multirow{2}{*}{\textbf{59.87}} & \multirow{2}{*}{\textbf{47.39}} & \multirow{2}{*}{\textbf{51.14}} & \multirow{2}{*}{\textbf{50.61}} \\
					~~~~+ Ref-NMS B & & & & & & & & \\
				\hline
			\end{tabular}
		} 
	\end{center}
	\caption{Performace (\%) on REC using Edgebox proposals.}
	\label{tab:edgebox-performance}
\end{table}
\addtolength{\tabcolsep}{3pt}

\section{Detailed Results on Recall of the Referent and Contextual Objects} \label{sec:s1}

The results on recall of the referent and contextual objects are reported in Table~\ref{tab:recall-of-referent} and Table~\ref{tab:recall-of-contextual}, respectively. For Table~\ref{tab:recall-of-referent}, as we can see, both two variants of Ref-NMS can help boost the recall of the referent on all dataset splits. More specifically, Ref-NMS B consistently achieves the best recall under all settings. Analogously, two variants can also improve the recall of contextual objects. However, for the contextual objects (\cf, Table~\ref{tab:recall-of-contextual}), it's hard to pick a winner between the two variants but it's safe to say that the best recall is always achieved by either of them. In conclusion, these results concretely validate the effectiveness of the proposed Ref-NMS in boosting the recall of the referent and contextual objects.

\addtolength{\tabcolsep}{-3pt}    
\begin{table*}[t]
	\small
	\begin{center}
		\scalebox{0.9}{
			\begin{tabular}{| l | c c c c c | c c c c c | c c c c c |}
				\hline
				\multirow{2}{*}{Models} & \multicolumn{5}{c|}{val} & \multicolumn{5}{c|}{testA} & \multicolumn{5}{c|}{testB}  \\

				& 0.5 & 0.6 & 0.7 & 0.8 & 0.9  & 0.5 & 0.6 & 0.7 & 0.8 & 0.9 & 0.5 & 0.6 & 0.7 & 0.8 & 0.9 \\
				\hline
					MAttNet~\cite{yu2018mattnet}$^\dagger$ &  75.49 & 72.80 & 67.95 & 54.92 & 16.83 & 79.46 & 77.71 & 72.69 & 58.85 & \textbf{13.79} & 68.62 & 64.99 & 59.76 & 48.71 & 20.82  \\
					~~+Ref-NMS B &  \textbf{77.16} & \textbf{74.27} & \textbf{69.00} & 55.74 & \textbf{17.24} & \textbf{80.78} & \textbf{78.72} & 73.50 & \textbf{58.94} & 13.72 & \textbf{71.78} & \textbf{68.24} & \textbf{62.26} & 49.81 & 21.28 \\
					~~+Ref-NMS R &  76.55 & 73.73 & 68.75 & \textbf{55.79} & 17.19 & 80.43 & 78.40 & \textbf{73.66} & 58.33 & 13.77 & 70.19 & 66.85 & 61.57 & \textbf{51.01} & \textbf{21.77} \\
				\hline
					NMTree~\cite{liu2019learning}$^\dagger$ &  75.15 & 72.45 & 67.58 & 54.45 & 16.59 & 79.58 & 77.73 & 72.74 & \textbf{58.87} & \textbf{13.88} & 68.64 & 65.10 & 60.02 & 49.17 & 21.35 \\
					~~+Ref-NMS B &  \textbf{77.13} & \textbf{74.04} & \textbf{68.61} & \textbf{55.53} & \textbf{16.95} & \textbf{80.22} & \textbf{78.03} & 72.95 & 58.62 & 13.77 & \textbf{71.64} & \textbf{67.99} & \textbf{61.79} & 49.72 & 21.43 \\
					~~+Ref-NMS R &  76.46 & 73.62 & 68.56 & 55.46 & 16.85 & 79.87 & 77.92 & \textbf{73.27} & 58.02 & 13.65 & 70.32 & 66.89 & 61.43 & \textbf{50.97} & \textbf{21.81} \\
				\hline
					CM-A-E~\cite{liu2019improving}$^\dagger$ &  76.75 & 73.98 & 69.06 & 55.65 & 16.93 & 81.23 & 79.28 & 74.30 & \textbf{60.19} & \textbf{14.21} & 70.09 & 66.38 & 60.92 & 49.66 & 21.33 \\
					~~+Ref-NMS B &  \textbf{78.91} & \textbf{75.70} & \textbf{70.31} & 56.56 & \textbf{17.34} & \textbf{82.11} & \textbf{80.01} & 74.95 & 60.12 & 14.05 & \textbf{73.66} & \textbf{70.05} & \textbf{63.69} & 51.17 & 21.88 \\
					~~+Ref-NMS R &  78.05 & 75.09 & 70.05 & \textbf{56.77} & 17.27 & 81.72 & 79.64 & \textbf{75.09} & 59.48 & 14.04 & 71.93 & 68.38 & 62.77 & \textbf{52.01} & \textbf{22.30} \\
				\hline
			\end{tabular}
		} 
	\end{center}
	\caption{RES Performance (\%) of different architectures in Pr@X metric on RefCOCO. $^\dagger$ denotes that the results are from our reimplementation using official released codes.}
	\label{tab:res-refcoco}
\end{table*}
\addtolength{\tabcolsep}{3pt} 

\addtolength{\tabcolsep}{-3pt}    
\begin{table*}[t]
	\small
	\begin{center}
		\scalebox{0.9}{
			\begin{tabular}{| l | c c c c c | c c c c c | c c c c c |}
				\hline
				\multirow{2}{*}{Models} & \multicolumn{5}{c|}{val} & \multicolumn{5}{c|}{testA} & \multicolumn{5}{c|}{testB}  \\
				
				& 0.5 & 0.6 & 0.7 & 0.8 & 0.9  & 0.5 & 0.6 & 0.7 & 0.8 & 0.9 & 0.5 & 0.6 & 0.7 & 0.8 & 0.9 \\
				\hline
				MAttNet~\cite{yu2018mattnet}$^\dagger$ &  64.71 & 62.44 & 58.54 & 47.76 & 14.33 & \textbf{70.12} & \textbf{68.44} & 63.87 & \textbf{51.89} & \textbf{12.29} & 56.00 & 52.85 & 48.27 & 39.23 & 17.24 \\
				~~+Ref-NMS B &  \textbf{65.16} & \textbf{62.82} & \textbf{58.74} & 47.69 & 14.37 & 69.61 & 67.78 & 63.41 & 50.38 & 12.03 & \textbf{57.13} & \textbf{53.63} & 48.89 & 39.17 & 16.55 \\
				~~+Ref-NMS R &  65.01 & 62.60 & 58.66 & \textbf{47.79} & \textbf{14.50} & 69.91 & 68.25 & \textbf{64.11} & 51.24 & 11.95 & 56.35 & 53.41 & \textbf{49.15} & \textbf{39.97} & \textbf{17.39} \\
				\hline
				NMTree~\cite{liu2019learning}$^\dagger$ &  65.24 & 62.92 & 58.98 & 47.56 & 14.34 & 69.89 & 68.11 & 63.64 & \textbf{52.03} & 12.38 & 56.49 & 53.32 & 49.21 & 39.33 & 17.30 \\
				~~+Ref-NMS B &  65.44 & 62.96 & 58.81 & 47.33 & 14.52 & 69.93 & 67.88 & 63.43 & 50.87 & \textbf{12.45} & 57.03 & 53.65 & 49.09 & 39.19 & 16.49 \\
				~~+Ref-NMS R &  \textbf{65.54} & \textbf{63.15} & \textbf{59.07} & \textbf{47.59} & \textbf{14.58} & \textbf{70.21} & \textbf{68.46} & \textbf{64.02} & 51.50 & 11.91 & \textbf{57.21} & \textbf{54.45} & \textbf{50.19} & \textbf{40.54} & \textbf{17.43} \\
				\hline
				CM-A-E~\cite{liu2019improving}$^\dagger$ &  66.98 & \textbf{64.60} & \textbf{60.63} & \textbf{49.15} & 14.73 & 71.46 & 69.70 & 65.14 & \textbf{53.23} & 12.31 & 57.13 & 53.92 & 49.34 & 40.17 & \textbf{17.55} \\
				~~+Ref-NMS B &  66.48 & 63.93 & 59.98 & 48.39 & 14.81 & \textbf{71.80} & 69.82 & 65.18 & 52.41 & \textbf{12.64} & \textbf{57.74} & \textbf{54.71} & \textbf{49.66} & 40.21 & 16.90 \\
				~~+Ref-NMS R &  \textbf{67.01} & 64.58 & 60.62 & 48.99 & \textbf{14.88} & 71.73 & \textbf{69.89} & \textbf{65.46} & 52.62 & 12.10 & 56.80 & 53.96 & 49.40 & \textbf{40.36} & 17.26 \\
				\hline
			\end{tabular}
		} 
	\end{center}
	\caption{RES Performance (\%) of different architectures in Pr@X metric on RefCOCO+. $^\dagger$ denotes that the results are from our reimplementation using official released codes.}
	\label{tab:res-refcoco+}
\end{table*}
\addtolength{\tabcolsep}{3pt} 

\addtolength{\tabcolsep}{-3pt}    
\begin{table*}[t]
	\small
	\begin{center}
		\scalebox{0.9}{
			\begin{tabular}{| l | c c c c c | c c c c c|}
				\hline
				\multirow{2}{*}{Models} & \multicolumn{5}{c|}{val} & \multicolumn{5}{c|}{test}  \\
				
				& 0.5 & 0.6 & 0.7 & 0.8 & 0.9  & 0.5 & 0.6 & 0.7 & 0.8 & 0.9 \\
				\hline
				MAttNet~\cite{yu2018mattnet}$^\dagger$ &  65.28 & 62.25 & 56.96 & 44.36 & 14.67 & 65.93 & 63.14 & 57.51 & 44.62 & 12.61 \\
				~~+Ref-NMS B &  66.03 & 62.70 & 57.09 & 44.34 & 14.73 & \textbf{66.58} & \textbf{63.49} & \textbf{57.82} & 45.21 & 13.03 \\
				~~+Ref-NMS R &  \textbf{66.26} & \textbf{63.38} & \textbf{57.74} & \textbf{44.93} & \textbf{15.09} & 65.86 & 62.83 & 57.24 & \textbf{45.26} & \textbf{13.12} \\
				\hline
				NMTree~\cite{liu2019learning}$^\dagger$ &  63.32 & 60.21 & 55.23 & 42.99 & 14.56 & 64.43 & 61.58 & 56.31 & 43.88 & 12.57 \\
				~~+Ref-NMS B &  \textbf{64.40} & 61.01 & 55.54 & 43.10 & 14.69 & \textbf{64.94} & \textbf{61.87} & \textbf{56.34} & 44.28 & 12.98 \\
				~~+Ref-NMS R &  64.01 & \textbf{61.07} & \textbf{55.86} & \textbf{43.67} & \textbf{15.20} & 64.54 & 61.41 & 56.09 & \textbf{44.62} & \textbf{13.25} \\
				\hline
				CM-A-E~\cite{liu2019improving}$^\dagger$ &  66.32 & 63.01 & 57.86 & 44.71 & 14.85 & 67.63 & 64.62 & 59.12 & 45.77 & 13.21 \\
				~~+Ref-NMS B &  \textbf{67.75} & \textbf{64.11} & \textbf{58.35} & \textbf{45.40} & 15.16 & \textbf{68.49} & \textbf{65.28} & \textbf{59.51} & 46.50 & 13.52 \\
				~~+Ref-NMS R &  66.52 & 63.52 & 58.03 & 45.04 & \textbf{15.28} & 68.29 & 65.06 & 59.48 & \textbf{46.96} & \textbf{13.82} \\
				\hline
			\end{tabular}
		} 
	\end{center}
	\caption{RES Performance (\%) of different architectures in Pr@X metric on RefCOCOg. $^\dagger$ denotes that the results are from our reimplementation using official released codes.}
	\label{tab:res-refcocog}
\end{table*}
\addtolength{\tabcolsep}{3pt} 

\section{Detailed Referring Expression Segmentation Results in Pr@X Metric} \label{sec:s2}

The detailed referring expression segmentation results with Pr@X metric on RefCOCO, RefCOCO+ and RefCOCOg are shown in Table~\ref{tab:res-refcoco}, Table~\ref{tab:res-refcoco+}, and Table~\ref{tab:res-refcocog}, respectively. We can observe that the Ref-NMS can consistently improve the grounding performance of all baselines and dataset splits over most of the metric thresholds. In particular, it is worth noting that the Ref-NMS significantly outperforms the baselines on RefCOCO testB split, RefCOCO+ testB split, where the category of the referent is more diverse, and on all splits of RefCOCOg, where the referring expressions are more complex. We attribute these performance gains to the richer contextual information conserved in the proposals generated by Ref-NMS, which is consistent with our motivation.

\section{Effectiveness of Pseudo Ground Truths} \label{sec:s3}

To validate the effectiveness of pseudo ground truths, we further design a strong baseline: Ref-NMS without pseudo ground truth. Both methods are trained using the same set of hyper-parameters and tested with CM-A-E. We can observe that Ref-NMS performs better on all splits but RefCOCO testB, where the performance difference between these two methods is trivial (0.05\%). These results validate the effectiveness of using pseudo ground truths for contextual objects in the training phase. Meanwhile, from another perspective, Ref-NMS trained with a single ground truth can be regarded as a specific one-stage referring expression comprehension model\footnote{Compared to the typical one-stage REC methods, this baseline is more lightweight and adaptable.}, which suggests that current one-stage REC methods are not qualified for generating proposals for two-stage REG methods.

\begin{table*}[h]
	\small
	\begin{center}
		\scalebox{0.9}{
			\begin{tabular}{| l | c c c | c c c | c c|}
				\hline
				\multirow{2}{*}{Model} & \multicolumn{3}{c|}{RefCOCO} & \multicolumn{3}{c|}{RefCOCO+} & \multicolumn{2}{c|}{RefCOCOg}  \\
				& val & testA & testB &  val & testA & testB & val & test  \\
				\hline
				CM-A-E & \multirow{2}{*}{80.56} & \multirow{2}{*}{83.63} & \multirow{2}{*}{\textbf{76.09}} & \multirow{2}{*}{67.33} & \multirow{2}{*}{71.83} & \multirow{2}{*}{58.64} & \multirow{2}{*}{69.14} & \multirow{2}{*}{68.76} \\
				~~~~+Ref-NMS w/o Pseudo GT &  & & & & & & &  \\
				CM-A-E & \multirow{2}{*}{\textbf{80.70}} & \multirow{2}{*}{\textbf{84.00}} & \multirow{2}{*}{76.04} & \multirow{2}{*}{\textbf{68.25}} & \multirow{2}{*}{\textbf{73.68}} & \multirow{2}{*}{\textbf{59.42}} & \multirow{2}{*}{\textbf{70.55}} & \multirow{2}{*}{\textbf{70.62}} \\
				~~~~+Ref-NMS & & & & & & & & \\
				\hline
			\end{tabular}
		} 
	\end{center}
	\caption{Performance (\%) on REC between baseline (\ie, Ref-NMS without pseudo ground-truths) and Ref-NMS.}
	\label{tab:multi-gt}
\end{table*}

\section{Performance using Edgebox Proposals Instead of Mask-RCNN Detector} \label{sec:s4}

To demonstrate the generalization ability of Ref-NMS, we carried out experiments using proposals from Edgebox~\cite{zitnick2014edge} instead of Mask-RCNN detector~\cite{he2017mask}. Specifically, at the first stage, we selected the top 8 scoring proposals according to either the Edgebox confidence (denoted as Edge-Conf) or the relatedness score of Ref-NMS. The recall of the referent are shown in Table~\ref{tab:edgebox-recall}. We further fed those proposals to the downstream MAttNet model and measured the performance on REC, which is shown in Table~\ref{tab:edgebox-performance}. We can observe that using relatedness score as a criterion to select proposals at the first stage can dramatically boost both the recall of referent and the performance on REC, which means the proposed Ref-NMS is not restricted to a specific object detector but, instead, can be applied to different kinds of proposals.

\begin{table*}[h]
	\small
	\begin{center}
		\scalebox{0.9}{
			\begin{tabular}{| c | c c c | c c c | c c|}
				\hline
				\multirow{2}{*}{Model} & \multicolumn{3}{c|}{RefCOCO} & \multicolumn{3}{c|}{RefCOCO+} & \multicolumn{2}{c|}{RefCOCOg}  \\
				& val & testA & testB &  val & testA & testB & val & test  \\
				\hline
					Edge-Conf & 14.99 & 15.10 & 14.98 & 15.17 & 14.79 & 15.83 & 16.16 & 15.21 \\
					Ref-NMS B & \textbf{85.62} & \textbf{89.11} & \textbf{81.35} & \textbf{86.44} & \textbf{88.25} & \textbf{80.71} & \textbf{82.82} & \textbf{83.08} \\
				\hline
			\end{tabular}
		} 
	\end{center}
	\caption{Recall (\%) of the referent using Edgebox proposals.}
	\label{tab:edgebox-recall}
\end{table*}

\section{More Qualitative Results on RefCOCO, RefCOCO+, and RefCOCOg} \label{sec:s5}

\begin{figure*}[t]
	\centering
	\includegraphics[width=1.0\linewidth]{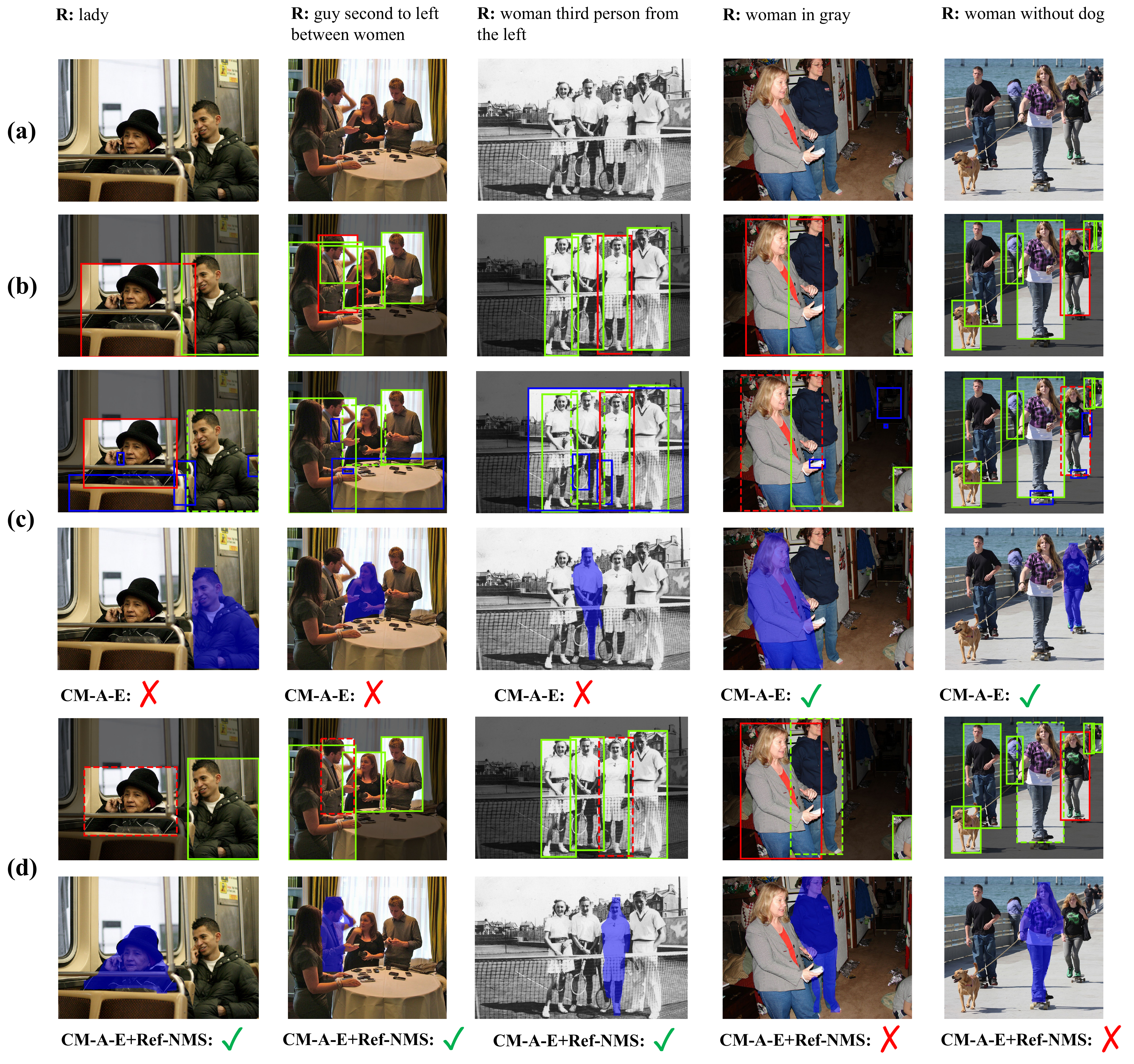}
	\caption{Qualitative REC and RES results on RefCOCO. \textbf{(a):} The referring expression and input image. \textbf{(b):} The annotated ground-truth bbox of the referent (marked in red) and the generated pseudo ground-truth bboxes of the contextual objects (marked in green). \textbf{(c):} The upper row demonstrates the proposals generated by an off-the-shelf object detector and the REC predictions of the downstream CM-A-E\cite{liu2019improving}; the lower row demonstrates the two-stage RES predictions acquired using the REC predictions from the upper row, the detailed method of which is fully described in~\cite{yu2018mattnet}. \textbf{(d):} Ref-NMS proposals, the REC and RES predictions from the downstream CM-A-E, arranged in the same format as \textbf{(c)}. The predicted bbox in REC is shown in dashed line. The denotations of bbox colors are as follows. \textbf{Red:} The bbox hits (IoU$>$0.5) the ground-truth bbox of the referent; \textbf{Green:} The bbox hits one of the pseudo ground-truth bboxes of the contextual objects; \textbf{Blue:} The false positive proposals.}
	\label{fig:viz_refcoco}
\end{figure*}

\begin{figure*}[t]
	\centering
	\includegraphics[width=1.0\linewidth]{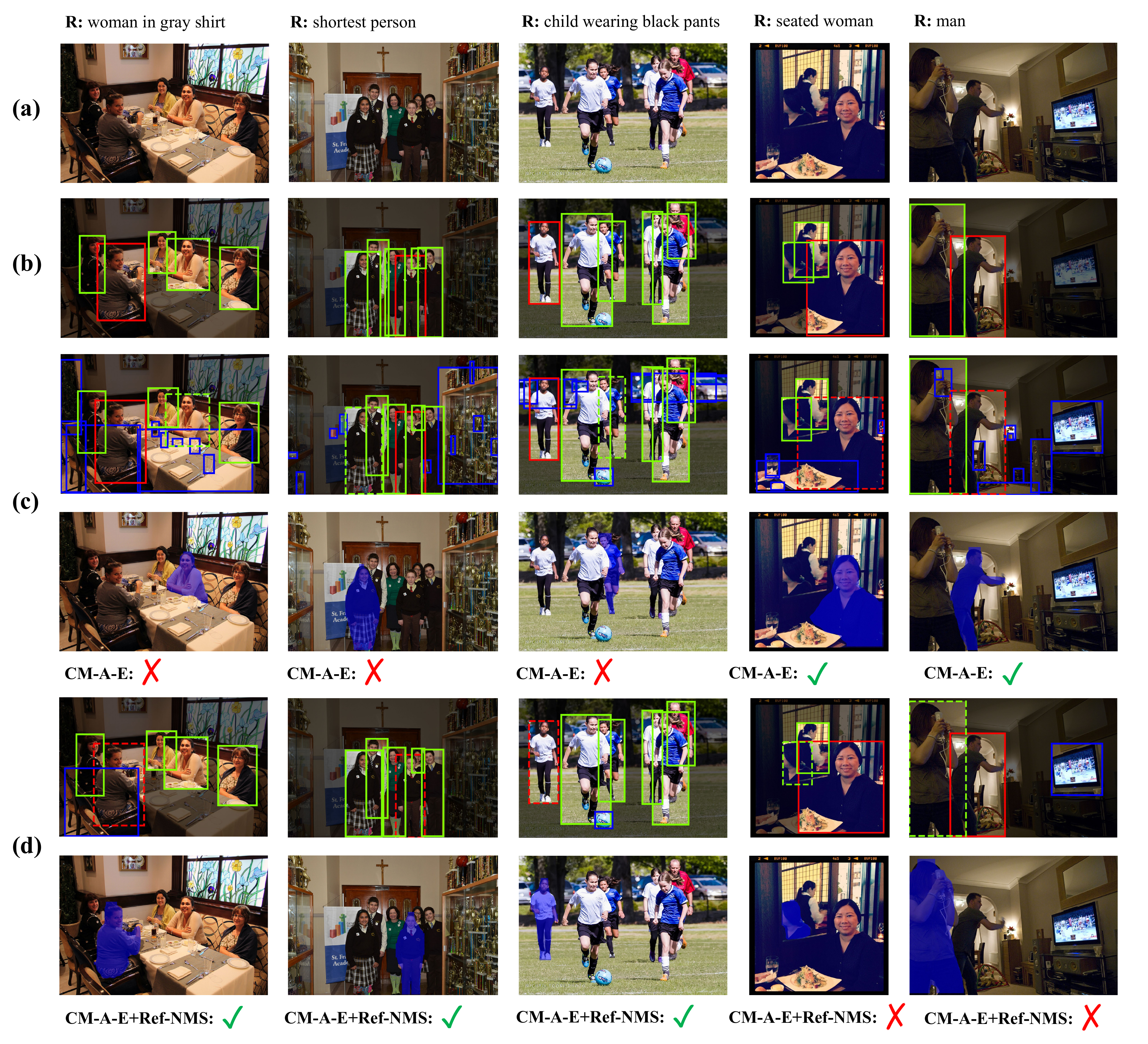}
	\caption{Qualitative REC and RES results on RefCOCO+, arranged in the same style as Figure~\ref{fig:viz_refcoco}.}
	\label{fig:viz_refcoco+}
\end{figure*}

\begin{figure*}[t]
	\centering
	\includegraphics[width=1.0\linewidth]{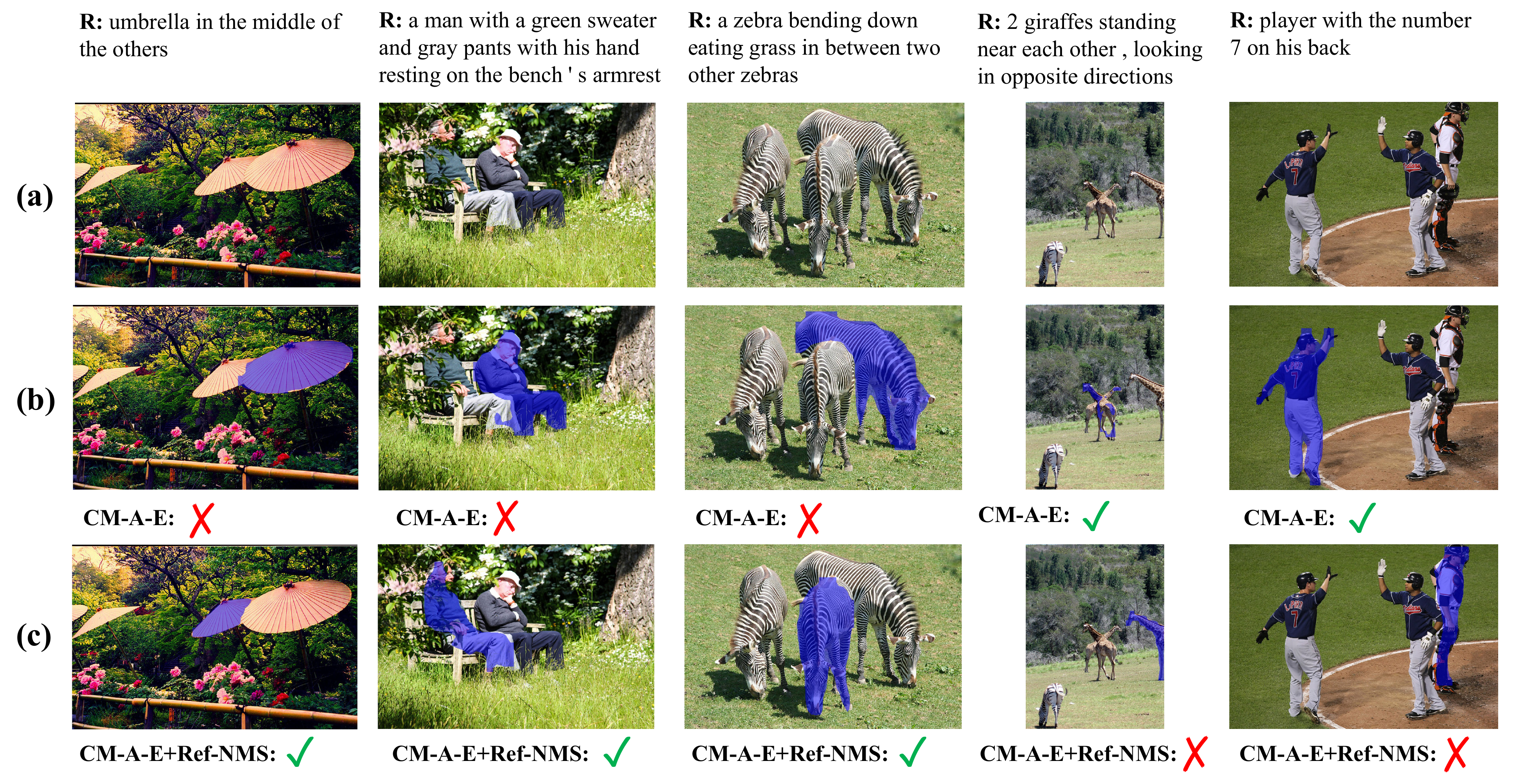}
	\caption{Qualitative RES results on RefCOCOg. The visualization of the proposals and REC results of the same examples used in Figure 5 in the main article.}
	\label{fig:viz_refcocog}
\end{figure*}

Qualitative results including the comparisons of proposals between the baseline (\ie, CM-A-E) and CM-A-E+Ref-NMS as well as the visualization of REC and RES predictions on RefCOCO and RefCOCO+ are illustrated in Figure~\ref{fig:viz_refcoco} and Figure~\ref{fig:viz_refcoco+}, respectively. The RES mask prediction results on RefCOCOg are shown in Figure~\ref{fig:viz_refcocog}.

From these qualitative results, we have the following observations: 1) The generated pseudo ground-truth bboxes can almost contain all contextual objects in the expression. 2) The baseline model tends to detect more false-positive proposals, and misses some critical objects. 3) Even in the failed cases on CM-A-E+Ref-NMS, Ref-NMS still generates express-aware proposals, and the grounding errors mainly come from the referent grounding step (\ie, the second stage).

\end{document}